\documentclass[IJPHM,2022]{PHMSociety}

\usepackage[T1]{fontenc}
\usepackage[utf8x]{inputenc}

\usepackage[scaled=.87]{helvet}

\usepackage{amsmath,amsthm,amssymb,amsfonts}
\usepackage{subcaption,graphicx}
\usepackage{xspace,xifthen}
\usepackage[dvipsnames,table]{xcolor}  
\usepackage{booktabs}
\usepackage{lipsum}
\usepackage[per-mode=symbol]{siunitx}  
\usepackage{relsize}
\usepackage{breakcites}
\usepackage{marginnote,tokcycle}
\usepackage[nameinlink]{cleveref}

\graphicspath{{imgs/},}  

\newboolean{tosubmit}
\setboolean{tosubmit}{true}

\newboolean{showhelpers}
\setboolean{showhelpers}{false}

\makeatletter
\renewcommand\paragraph{\@startsection{paragraph}{4}{\z@}%
   {.7ex \@plus.4ex \@minus.2ex}%
   {-.8em \@plus-.2em \@minus-.1em}%
   {\normalfont\normalsize\bfseries}}
\makeatother

\makeatletter
\newcommand{\nl@ref}[1]{%
  \cref@gettype{#1}{\@temptype}%
  \cref@getcounter{#1}{\@tempctr}%
  \def\nlt{\the\csname c@\@temptype\endcsname}%
  \ifnum\nlt=\numexpr\@tempctr-1\relax%
    \nlcase the next \namecref{#1}
  \else%
  \ifnum\nlt=\numexpr\@tempctr+1\relax%
    \nlcase the previous \namecref{#1}
  \else%
    \cref{#1}%
  \fi\fi
}
\newcommand{\nlref}{\let\nlcase\relax\nl@ref}
\newcommand{\Nlref}{\let\nlcase\MakeUppercase\nl@ref}
\makeatother

\Crefname{equation}{Eq.}{Eqs.}
\crefname{equation}{equation}{equations}
\Crefname{figure}{Fig.}{Figs.}
\crefname{figure}{figure}{figures}
\Crefname{tabular}{Table}{Tables}
\crefname{tabular}{table}{tables}
\Crefname{definition}{Def.}{Defs.}
\crefname{definition}{definition}{definitions}
\Crefname{proposition}{Prop.}{Props.}
\crefname{proposition}{proposition}{propositions}
\Crefname{section}{Sec.}{Secs.}
\crefname{section}{section}{sections}

\ifthenelse{\boolean{showhelpers}}{%
  \usepackage[pagewise]{lineno}  
  \linenumbers
  
  \setlength\linenumbersep{.5em}
}{}

\ifthenelse{\boolean{tosubmit}}{}{ 
  \usepackage[firstpage]{draftwatermark}  
  \SetWatermarkColor{Blue!11}
}

\makeatletter
\def\THICKhrulefill{\leavevmode \leaders \hrule height 5pt\hfill \kern \z@}
\def\getfirst#1#2\relax{\tctestifnum{\count@stringtoks{#1}>1}{ERROR}{#1}}
\makeatother
\makeatletter  
\let\oldmarginnote\marginnote
\renewcommand*{\marginnote}[1]{%
   \begingroup%
   \ifodd\value{page}
	 \if@firstcolumn\else\reversemarginpar\fi
   \else
	 \if@firstcolumn\reversemarginpar\fi
   \fi
   \oldmarginnote{#1}%
   \endgroup%
}
\makeatother
\def\MARGINALNOTEWIDTH{.5\linewidth}
\newcommand{\colorpar}[3]{\colorbox{#1}{\parbox{#2}{#3}}}
\newcommand{\marginremark}[3]{%
  \ifthenelse{\boolean{tosubmit}}{}{
	\marginnote{\hspace*{-1.5em}\colorpar{#2}{\MARGINALNOTEWIDTH}{\color{#1}#3}}
}}
\newcommand{\highlightedremark}[4]{%
  \ifthenelse{\boolean{tosubmit}}{}{
	\begin{center}\fcolorbox{#1}{#2}{%
	\begin{minipage}{.98\linewidth}\color{#1}%
	\textbf{\THICKhrulefill[ #3 ]\THICKhrulefill}%
	\par\noindent#4\end{minipage}}\end{center}%
}}
\newcommand{\hey}[4]{%
  \ifthenelse{\boolean{tosubmit}}{}{
  \leavevmode\marginremark{#1}{white}{\sffamily\large@\getfirst#3\relax\relax}
  \colorbox{#2}{\sffamily\bfseries{@#3:}}~{\sffamily\color{#1}#4}}}
\newcommand{\todo}[1]{%
  \ifthenelse{\boolean{tosubmit}}{}{
  \noindent\textsf{\color{Red}\textbf{TODO:} #1}%
  \marginremark{red}{white}{\textsf{\bfseries TODO}}}}
\newcommand{\tocite}[1][??]{%
  \ifthenelse{\boolean{tosubmit}}{}{
  \noindent\textbf{\sffamily\textcolor{blue!85}{[#1]}}%
  \marginremark{blue}{white}{\textsf{\bfseries CITE!}}}}
\colorlet{DB-fg}{BrickRed!40!red}
\colorlet{DB-bg}{Peach!30}
\colorlet{CEB-fg}{TealBlue!75!green!75!black}
\colorlet{CEB-bg}{Aquamarine!8}
\colorlet{shade}{gray!13}
\newcommand{\hours}[1][]{\ensuremath{\text{\if\relax\detokenize{#1}\unit{\hour}\relax\else\qty{#1}{\hour}\fi}}\xspace}
\newcommand{\Celsius}[1][]{\ensuremath{\text{\if\relax\detokenize{#1}\unit{\degreeCelsius}\relax\else\qty{#1}{\degreeCelsius}\fi}}\xspace}
\newcommand{\card}[1]{\ensuremath{\left|{#1}\right|}\xspace}
\newcommand{\gain}[1][\theta]{\ensuremath{G^{#1\!}}\xspace}

\newcommand{\ms}{\scriptscriptstyle} 

\newcommand{\leftSplit}{\ensuremath{D^{\theta}_{\ms \leqslant}}\xspace}
\newcommand{\rightSplit}{\ensuremath{D^{\theta}_{\ms >}}\xspace}

\begin{document}

\title{Automated fault tree learning from continuous-valued sensor data:\\a case study on domestic heaters}

\author{%
 	Bart Verkuil\authorNumber{1},
	Carlos E.\ Budde\authorNumber{2},
	Doina Bucur\authorNumber{3}
}

\address{
	\affiliation{{1,3}}{Data Management \& Biometrics, University of Twente, Enschede, 7522~NB, The Netherlands}{
		{\email{d.bucur@utwente.nl}}
	}
	\tabularnewline
	\affiliation{2}{Security, University of Trento, Trento, I-38122, Italy}{
		{\email{carlosesteban.budde@unitn.it}}
	}
}

\maketitle
\pagestyle{fancy}
\thispagestyle{plain}

\phmLicenseFootnote{Bart Verkuil} 

\begin{abstract}
Many industrial sectors have been collecting big sensor data.
With recent technologies for processing big data, companies can exploit this for automatic failure detection and prevention.
We propose the first completely automated method for failure analysis, machine-learning fault trees from raw observational data with continuous variables.
Our method scales well and is tested on a real-world, five-year dataset of domestic heater operations in The Netherlands, with 31 million unique heater-day readings, each containing 27 sensor and 11 failure variables.
Our method builds on two previous procedures: the C4.5 decision-tree learning algorithm, and the LIFT fault tree learning algorithm from Boolean data.
C4.5 pre-processes each continuous variable: it learns an optimal numerical threshold which distinguishes between faulty and normal operation of the top-level system.
These thresholds discretise the variables, thus allowing LIFT to learn fault trees which model the root failure mechanisms of the system and are explainable.
We obtain fault trees for the 11 failure variables, and evaluate them in two ways: quantitatively, with a significance score, and qualitatively, with domain specialists. Some of the fault trees learnt have almost maximum significance (above 0.95), while others have medium-to-low significance (around 0.30), reflecting the difficulty of learning from big, noisy,  real-world sensor data. The domain specialists confirm that the fault trees model meaningful relationships among the variables. 
\end{abstract}

\section{Introduction}
\label{sec:intro}



Fault tree analysis is a world-leading standard for safety and reliability assessment \cite{RS15,VSD+02}, with growing applications in emergent fields such as cybersecurity \cite{NFW17}.

Fault trees are logical models for the propagation of basic functional failures.
From an engineering perspective, a fault tree (FT) is a graphical representation of the possible failure modes of a system, i.e.\ the distinct observable failure processes of system functions, broken down into intermediate failures and their interactions \cite{RBN+19}.

FT models are ubiquitous for reliability, availability, maintainability, and safety (RAMS) analyses as they are interpretable by human experts, and extremely versatile: an FT can describe the failure behaviour of a system or process to almost any required level of detail.
However, the traditional expert-driven FT building practice is resource intensive, subjective, and error-prone \cite{RS15}.
As a result, engineers and researchers are looking for ways to automate this modelling step \cite{NBS18,LNK20,TBB+20}.


\subsection{Building FTs directly from data}
\label{sec:intro:goal}

The advent of the big data era has opened new possibilities for the automatic construction of FT models.
At the same time, it exacerbates the combinatorial explosion in the number of models that can be derived.
This is inherent to the structure of an FT%
, 
where each data variable is a potential leaf or gate.

\begin{figure}[ht]
	\centering
	\includegraphics[width=.8\linewidth]{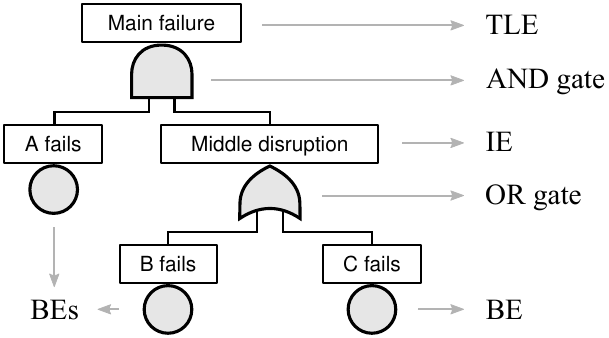}
	\caption{\textbf{Fault tree model} structure and terminology}
	\label{fig:sample_FT}
\end{figure}

\paragraph{Fault Tree Models.}
\!\!Technically, an FT is a single-rooted directed acyclic graph (DAG), whose leaves are called basic events (BEs) and represent indivisible failures in the system, such as ``no power input'' or ``insufficient water supply''.
These BEs are connected to intermediate event nodes (IE): when the failure corresponding to a BE takes place, it propagates towards the connected IEs.
In turn, each intermediate event has an output that can be connected as input to an upper IE, thus generating the DAG structure---see \Cref{fig:sample_FT}.

When the inputs of an intermediate event receive a fail signal, the IE can propagate this failure to its output.
The propagation mechanism is defined by a logical gate that labels the IE:
OR gates propagate a failure if any input fails;
AND gates only do it when all its inputs fail.
There are further gates such as VOT: for an in-depth description of FTs we refer the interested reader to \cite{RS15}.

The root of the tree is identified with the node at its top, called the top level event (TLE).
The TLE represents the main event of interest that the FT models, e.g.\ a heater system failure.
In turn, the BEs represent elemental failures that can lead to such an event, e.g.\ ``insufficient water supply''.
The FT is said to fail, when the failures of certain BEs propagate and cause failures that eventually reach the TLE.

\paragraph{FT construction.}
Traditionally, building an FT is a human-consensus process that includes experts from the areas related to the corresponding asset/system \cite{VSD+02,Ber09}.
The process starts with the identification of the TLE, which defines the type of failure to study.
This is a non-trivial initial decision: compare e.g.\ ``heater does not start'' vs.\ ``heater stops after short operation''.
Both involve a heater system failure, but the causes leading to each type of failure can be completely different, and this impacts the resulting FTs \cite{RS15}.

Once the TLE has been agreed upon, a functional analysis is carried out to determine the relevant causes that could lead to it.
Then the causes of these causes are investigated and so on, until failures that are deemed elemental (the BEs) are reached.
All of this involves expert knowledge, technical manuals, revisions, and discussions that span for weeks or months.

\paragraph{Big data.}
However, increasingly many assets are manufactured to record elemental failures (Booleans) and performance indicators (continuous sensor readings);
and ever since the Internet of Things began, these readings are poured into the data streams of most companies \cite{Eva11}.

This brings a new approach to risk analysis, where data comes first and (some) failures can be found based on it.
More precisely, modern companies can take advantage of the data already available to them, determining its failure-predictive capabilities.
However and unlike FTs, the learning models used to find correlations between failures and data are not necessarily human-intelligible.
This hinders an adoption of the big data solution by many industrial sectors, specially those that traditionally use conservative risk management processes, e.g.\ aviation, nuclear power, housing, and railway operation \cite{CHM07,Gri16}. 

Coincidentally, such industries are among the main users of FTs for safety analysis, given the versatility and high degree of explainability of FTs.
These industries, and any companies with online readings of sensor and failure data, can greatly benefit by deriving FTs automatically from such data.
That is the main contribution of this study: \emph{a scalable learning algorithm to learn FTs from continuous and Boolean data, with a demonstration on the assets of a company that operates over 30 million domestic heaters across The Netherlands}.

\paragraph{Prior work.}
There is recent work on deriving FTs from failure data.
This data can be obtained auto\-mat\-ic\-ally---via online data streams---or semi-automatically---recorded in situ and then collected. 
In either case and to the best of our knowledge, all literature assumes a Boolean data input, i.e.\ true/false indicators of failures in the components of the main asset \cite{RS15,NBS18,LBS19,LNK20}.

In contrast, our approach can process \emph{continuous-valued data} coming from performance indicators, viz.\ sensors readings, which we relate to failure behaviour by learning a failure threshold per sensor. We use the C4.5 learning algorithm for this \cite{Qui93}, originally designed for decision trees---another type of tree model which can visualise the hierarchical propagation of any decisions, including failure. 

Decision trees are in some respects simpler than FTs: failure propagation in decision trees is not regulated by logical gates.
However, in other respects the decision trees are richer in possibilities: they are learnt from continuous-valued data, and this can be done automatically, detecting thresholds for each variable such that a value below the threshold models a behaviour---e.g.\ failure---that differs from the behaviour when the value is above the threshold.

The connection between FTs and decision trees has been used before, for instance in \cite{LAR05} to derive the latter from existing FT models.
Here we exploit the opposite direction, making use of the advantages that decision trees bring, to provide a realistic integrated method to learn FTs from real-world data.
In that sense our work is closer in methodology to e.g.\ \cite{ADM+18}, where the main goal in our case is to use FT models to describe the governing fault mechanisms found.

Besides C4.5, our work builds on the LIFT algorithm \cite{NBS18}, which we modify to permit variable dataset sizes, and then use to learn the FT structure that fits best the failure under analysis.
For this, LIFT employs the Mantel-Haenszel statistical test to correlate events.
Other probabilistic causal models could be used, such as Bayesian Networks, to learn relations between Boolean events and component failures.
This has been studied in \cite{LBS19,LBBS19}, and applied to synthetic benchmarks, and failures in printer nozzles---all starting from Boolean input.
However, using Boolean data (only) when continuous data is also be available, fails to capture failure mechanisms that surface in e.g.\ sensor readings \cite{TBB+20}.
In view of this, our work presents a step towards more general, automatic learning algorithms.

\subsection{Goal and approach}
\label{research_questions}

The main goal of this work is to use continuous sensor data collected by existent company assets, to build FTs for the failure mechanism of these assets---inasmuch as these failures are correlated to the data---in an automatic and scalable manner.
To orient our investigation, we formulate this goal in term of the following research questions:
\begin{enumerate}
\itemindent2em
\renewcommand{\theenumi}{RQ\arabic{enumi}}
\renewcommand{\labelenumi}{\textbf{\theenumi}:}
\item	\label{RQ:howto}
		Which automatic method(s) can generate
		a structure of failure mechanisms,
		conforming to the FT standard,
		from a continuous-valued data input of sensor readings?
\item	\label{RQ:howgood}
		How to measure the quality of the resulting FTs, in terms of the
		correlation between the sensors data and the failures that they
		are expected to predict?
\item   \label{RQ:big}
        Is the learning method directly applicable to
		big industrial input data?
\end{enumerate}

In \ref{RQ:howto}, ``automatic methods'' stands for algorithms that do not require human input, and can learn the FTs solely from the sensor data fed into the learning procedure.

In \ref{RQ:howgood}, we are mostly interested in quantitative quality metrics.
Nevertheless, we also include the qualitative aspect of \emph{explainability}: these are enquiries to experts (in the application domain of the case study) on whether the resulting FT structures match their intuition, or if they are instead difficult to comprehend from their expert perspective.

For \ref{RQ:big}, we ask whether the runtime of this learning method scales with large data sizes, particularly when there are many failure notifications in the dataset to learn from.

\paragraph{Outline and main results.}
We implemented our results in an industrial case study, described next in \Cref{sec:case_study}.
The details of our FT learning procedure are given in \Cref{sec:method}, followed in \Cref{sec:results} by the results of its application to the case study.

Regarding \ref{RQ:howto}, we found that \emph{feeding continuous data to the C4.5 learning algorithm}---to interpret Boolean failure behaviour out of the sensor readings---\emph{and then passing the results to LIFT}---to learn FT structures from the Boolean-in\-ter\-preted data---\emph{can generate FTs that relate continuous values to explainable failure mechanisms}.

As for \ref{RQ:howgood}, we used the concept of gate significance introduced in \cite{NBS18}, but replacing the PAMH score by the phi coefficient to allow for changes in the data size and dimensionality \cite{Cra46}.
\emph{The quality of an FT is then measured as the significance of its top gate, interpreted as a lower bound on the correlation between gate inputs and output}, across all the gates of the FT.

The application of this approach to our case study of domestic heaters resulted in 44 FTs.
Of these, the FTs of failures related to the supply- or return-water temperature achieved significance as high as $0.96$ (higher is better, the range of values is $[-1.0,1.0]$).
For almost every other failure, an FT was produced with significance around $0.3$.
Moreover, in cases where expert technicians had good understanding of the underlying failure mechanisms, they confirmed that the relevant sensor variables were included in the corresponding FTs.

The above also answers \ref{RQ:big} positively:
\emph{we could successfully work with sensor data, recorded daily between 2015 and 2020 by Dutch heaters, totalling 31 million unique heater-day readings}.
Per fault tree, the runtime of our learning algorithm in a standard desktop computer is in the order of minutes.

\section{The case study}
\label{sec:case_study}

We analyse a large dataset containing time series of con\-tinu\-ous-\-valued sensor data, collected automatically by Intergas Heating.
The data describes the operation of Intergas domestic heaters---the Combi Compact HRE models---installed at consumers' houses throughout The Netherlands \cite{Int18}.
Between 2015 and 2020, data sensed locally by each connected heater was collected every \hours[24].
The resulting dataset consists of multiple terabytes of raw data, with 31 million unique heater-day data collections, each with data from 27 built-in sensors and 11 self-diagnosed system failures.

\begin{table}[htb]
  \caption{\textbf{Heater sensor variables} (27 in total)}
  \label{tab:dataset}
  \centering
  \begingroup
  \smaller
  \rowcolors{2}{shade}{white}
  \renewcommand{\arraystretch}{1.1}
  \def\units#1{\ensuremath{\left(\unit{#1}\right)}}
  \begin{tabular}{l@{~~~}p{.62\linewidth}}
	\toprule
	\textbf{Sensor variable}	&	\textbf{Description} \\
	\midrule
	bc\_tapflow				&	Water usage
	                            \units{\litre\per\minute} \\
	boilertemp				&	Machine temperature
	                            \units{\degreeCelsius} \\
	burnerstarts\_24h		&	Number of starts per \hours[24] \\
	ch\_pressure			&	Water pressure
	                            \units{\bar} \\
	flue\_sided\_resistance	&	Torque for fan rotation in flue duct
	                            \units{\newton\metre} \\
	gasmeter\_ch\_24h		&	Gas used per \hours[24] for central heating
	                            \units{\cubic\metre} \\
	gasmeter\_dhw\_24h		&	Gas used per \hours[24] for domestic hot water
	                            \units{\cubic\metre} \\
	heaterload\_ch\_24h		&	Delivered power \% of full power over last \hours[24],
	                            per hour, in central heating mode (\%) \\
	heaterload\_ch\_total	&	Delivered power \% of full power over last \hours[24],
	                            total, in central heating mode (\%) \\
	heaterload\_dhw\_24h	&	Delivered power \% of full power over last \hours[24],
	                            per hour, in domestic hot water mode (\%) \\
	heaterload\_dhw\_total	&	Delivered power \% of full power over last \hours[24],
	                            total, in domestic hot water mode (\%) \\
	heatertemp				&	Boiler temperature
	                            \units{\degreeCelsius} \\
	io\_curr\_high			&	Ionization current on high output power
	                            \units{\micro\ampere} \\
	io\_curr\_low			&	Ionization current on low output power
	                            \units{\micro\ampere} \\
	outside\_temp			&	Outside temperature
	                            \units{\degreeCelsius} \\
	override\_outside\_temp	&	Alternative temperature measure
	                            \units{\degreeCelsius} \\
	pump\_pwm				&	Pulse-width modulation control signal for modulating pump
	                            (\%) \\
	room\_override\_zone1	&	Override temperature for zone 1
	                            \units{\degreeCelsius} \\
	room\_override\_zone2	&	Override temperature for zone 2
	                            \units{\degreeCelsius} \\
	room\_set\_zone1		&	Set temperature for zone 1
	                            \units{\degreeCelsius} \\
	room\_set\_zone2		&	Set temperature for zone 2
	                            \units{\degreeCelsius} \\
	room\_temp\_zone1		&	Measured temperature for zone 1
	                            \units{\degreeCelsius} \\
	room\_temp\_zone2		&	Measured temperature for zone 2
	                            \units{\degreeCelsius} \\
	s1\_temp				&	Supply water temperature
	                            \units{\degreeCelsius} \\
	s2\_temp				&	Return water temperature
	                            \units{\degreeCelsius} \\
	s3\_temp				&	Warm water temperature
	                            \units{\degreeCelsius} \\
	waterflow\_ch			&	Water used for central heating
	                            \units{\litre} \\
	\bottomrule
  \end{tabular}
  \endgroup
\end{table}


\paragraph{Sensor variables.}
The 27 built-in sensors measure internal variables, such as water usage or the temperature at various zones, and an external variable, namely outside temperature. 
All variables are listed in alphabetical order in \Cref{tab:dataset}, with a description in terms of components of the Combi Compact HRE service instructions \cite{Int18}.
For each variable, four basic statistics are recorded or computed automatically every day: the minimum, maximum, average, and range. 

\paragraph{Failure variables.}
The values of all sensors are monitored by a control loop running in the heater.
Some sensors perform safety-critical measurements.
For example, sensors s1\_temp and s2\_temp from \Cref{tab:dataset} are located in the heat exchanger, and measure whether the heat from the gas is transferred to the water at the expected rate.
If a critical sensor reports values surpassing a threshold, the heater diagnoses this as a \emph{system failure} (a Boolean), and marks the failure in the data collected for the day.
We study 11 such Boolean failures, listed in \Cref{tab:failures}, which are present in the input data.
The failure variables ``Warning low t1'' and ``Warning low t2'' are triggered by sensors s1\_temp and s2\_temp, so they are expected to be highly correlated.
Similarly, the failure variable ``Warning low ch\_pressure'' reports low water pressure.
Depending on the type of heater, a normal pressure is between 1 and \SI{1.5}{\bar}; by the company's business model, a value of \SI{0.49}{\bar} triggers a failure diagnosis.
No other failure variable is directly associated to a sensor variable from \Cref{tab:dataset}.

\begin{table}[htb]
  \caption{\textbf{Boolean failure modes} (11 in total)}
  \label{tab:failures}
  \centering
  \begingroup
  \smaller
  \rowcolors{2}{shade}{white}
  \renewcommand{\arraystretch}{1.1}
  \begin{tabular}{l p{40mm}}
	\toprule
	\textbf{Failure}			&	\textbf{Description} \\
	\midrule
	Lockout code 0 				&	Sensor fault at self check \\
	Lockout code 4 				&	No flame signal \\
	Lockout code 5 				&	Poor flame signal \\
	Lockout code 8 				&	Incorrect fan speed \\
	Lockout code 11 			&	Fault in S1 flow (vent) \\ 
	Lockout code 13 			&	Fault in S1 flow (switch) \\ 
	Warning flame lost			&	Flame lost \\
	Warning ignition failed		&	Ignition failed 4 times \\
	Warning low ch\_pressure	&	Channel pressure below \qty{0.49}{\bar} \\
	Warning low t1				&	Temperature s1\_temp at \Celsius[0] \\
	Warning low t2	 			&	Temperature s2\_temp at \Celsius[0] \\
	\bottomrule
  \end{tabular}
  \endgroup
\end{table}

In summary, the table of available data consists of 27 real-valued sensor columns---each with four daily statistics computed over
high-frequency
readings---and 11 Boolean failure columns.
Each row in this data table corresponds to the readings of a heater during a day of operation.
We analyse each failure mode independently, using all sensor variables and data collected from all the heaters which exhibited such a failure, to gain a general understanding of how raw sensor readings are associated to each system failure.

\paragraph{Data quality.}
We remove duplicates from the raw dataset, as well as data from heaters with clearly corrupt readings, e.g.\ with values out of feasible physical ranges.
Despite this preprocessing, a large amount of noise remains in the data due to individual sensors with temporal malfunction or disconnections: these can report values near the minimum or maximum range, or the value zero.
For example, some temperature sensors occasionally report the value \Celsius[327.67], which is the maximum recordable temperature%
\footnote{%
	This is due to the use of 16-bit registers to hold signed integers,
	which results in a maximum value of $2^{15}-1=32767$.}\!.
Since they occur in real data, we choose to preserve these extreme readings in the dataset, and learn from them.
This data-cleaning process is, as usual, heavily dependent on the nature of the data; as such, we do not include it as part of our proposed solution to the research questions.

\begin{figure}[htb]
    \centering
    \includegraphics[width=.95\linewidth]{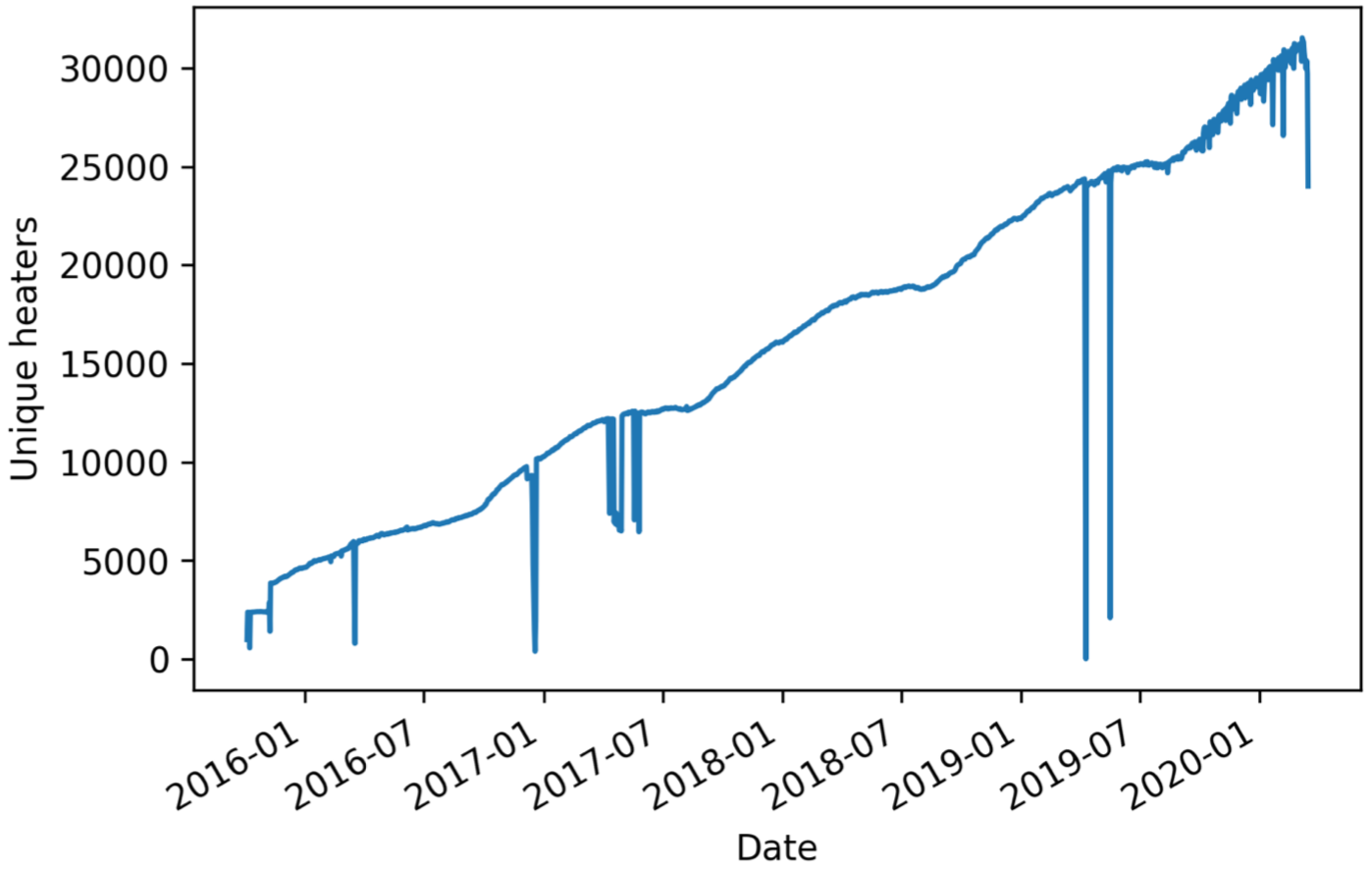}
    \caption{\textbf{Number of unique heaters recorded per day}.
	        Drops in the curve indicate missing values,
	        due to malfunctioning heaters or network communication issues.}
    \label{fig:data_collection}
\end{figure}

There are also missing readings, caused either by a failed heater or by failed network communication during data collection%
\footnote{%
	The latter is explained by the use of the UDP transport protocol, 
	which does not guarantee delivery nor duplicate protection.}\!.
\Cref{fig:data_collection} shows the number of (unique) heater readings recorded per day, from the beginning of data recordings at the company in 2015.
The positive slope of the curve reflects the installation of new heaters by Intergas across The Netherlands.
The drops in the line indicate missing data: while roughly 90\% of the heaters on record miss at least one reading during this five-year period, data overall remains abundant.

\section{Method}
\label{sec:method}

The data from the previous \namecref{sec:case_study} is automatically collected by the Combi Compact HRE heaters of Intergas.
To address \ref{RQ:howto} we design an algorithm that uses it to learn a fault tree for each system failure, without human input in the process.

Since the variables from \Cref{tab:failures} were designed as Boolean indicators of heater failure, we use them as TLEs of our fault trees.
Below these TLEs, the intermediate and basic events will be formed with automatically-selected sensor variables, which are found to be correlated to the top event via the methods detailed in \Cref{sec:method:thresholds,sec:method:learning}.
As a result, the gates of the FT link the behaviour of the sensors to the TLE, modelling how failures (are learnt to) propagate across system subcomponents.
However, fault trees encode Boolean relations, while our sensor readings are real-valued variables.
We bridge this gap by discretising each sensor variable.

\subsection{Learning failure thresholds for sensor variables}
\label{sec:method:thresholds}

The first step of our approach is to learn real-valued \emph{thresholds} for the sensor variables, such as \Celsius[27.21] for the minimum daily value of the temperature variable s2\_temp from \Cref{tab:dataset}.
Each threshold applies only in the context of a particular failure mode: for instance the previous example is related to the failure variable ``Lockout code 4''.
In contrast, the threshold \Celsius[38.81] was found for the range of daily values of s2\_temp in relation to the failure variable ``Lockout code 11''.

To evaluate how good a threshold $\theta\in\mathbb{R}$ is, we use a metric based on information gain for Boolean variables.
Essentially, the gain measures the correlation between the failure variable, and the sensor variable split by the threshold $\theta$.
This follows the internal logic of the C4.5 algorithm \cite{Qui93}. 
We now explain in detail how we use it for our purposes.

An optimal threshold is learnt for each combination of sensor variable, statistic, and failure variable.
Let $s$ denote the statistic of a sensor variable (e.g.\ minimum daily value of s2\_temp), and $f$ a failure variable (e.g.\ ``Warning low t1'').
These are two columns in our data table: $s$ is a real-valued column, and $f$ a Boolean-valued column.
In the corresponding dataset $D=s\mid f$, let $p_0$ denote the proportion of class 0 in column $f$: this represents normal operation, viz.\ absence of failure recordings.
In turn, let $p_1$ denote class 1 (failure).
The \emph{information entropy} $E$ is then defined as a weighted sum of these class probabilities, which reaches value 1 if the classes are equal in size, and value 0 if one class is empty:
\[
	E(D) = - \big(p_0 \cdot \log(p_0) + p_1 \cdot \log(p_1)\big) .
\]

Then, the dataset $D=s\mid f$ is split by a proposed threshold $\theta\in\mathbb{R}$ into a left and a right split.
The left split, denoted $\leftSplit\subseteq D$, are the rows where the sensor values $s$ are lower than $\theta$.
The right split $\rightSplit\subseteq D$ is analogously defined.

\paragraph{The gain of a threshold.}
The \emph{gain} $\gain(D)$ of a proposed threshold $\theta$ for the dataset $D$ is defined as the difference in entropy between the unsplit dataset, and the weighted sum of the entropy values after the split:
\[
	\gain(D) = E(D)
		- \frac{\card{\leftSplit}} {\card{D}} E\big(\leftSplit\big)
		- \frac{\card{\rightSplit}}{\card{D}} E\big(\rightSplit\big) ,
\]
where $|D|$ is the size in number of rows of dataset $D$, so the weights are the fractions of readings on each side of the split.

Intuitively, \gain measures the amount of entropy removed by the split at $\theta$.
In a perfect split, all sensor values $s$ below $\theta$ would be part of one class of $f$ (either normal or failure), and the remaining sensor values would be part of the other class.
More generally, the better the threshold, the higher the gain.

\begin{figure}[ht]
    \centering
    \includegraphics[width=.95\linewidth]{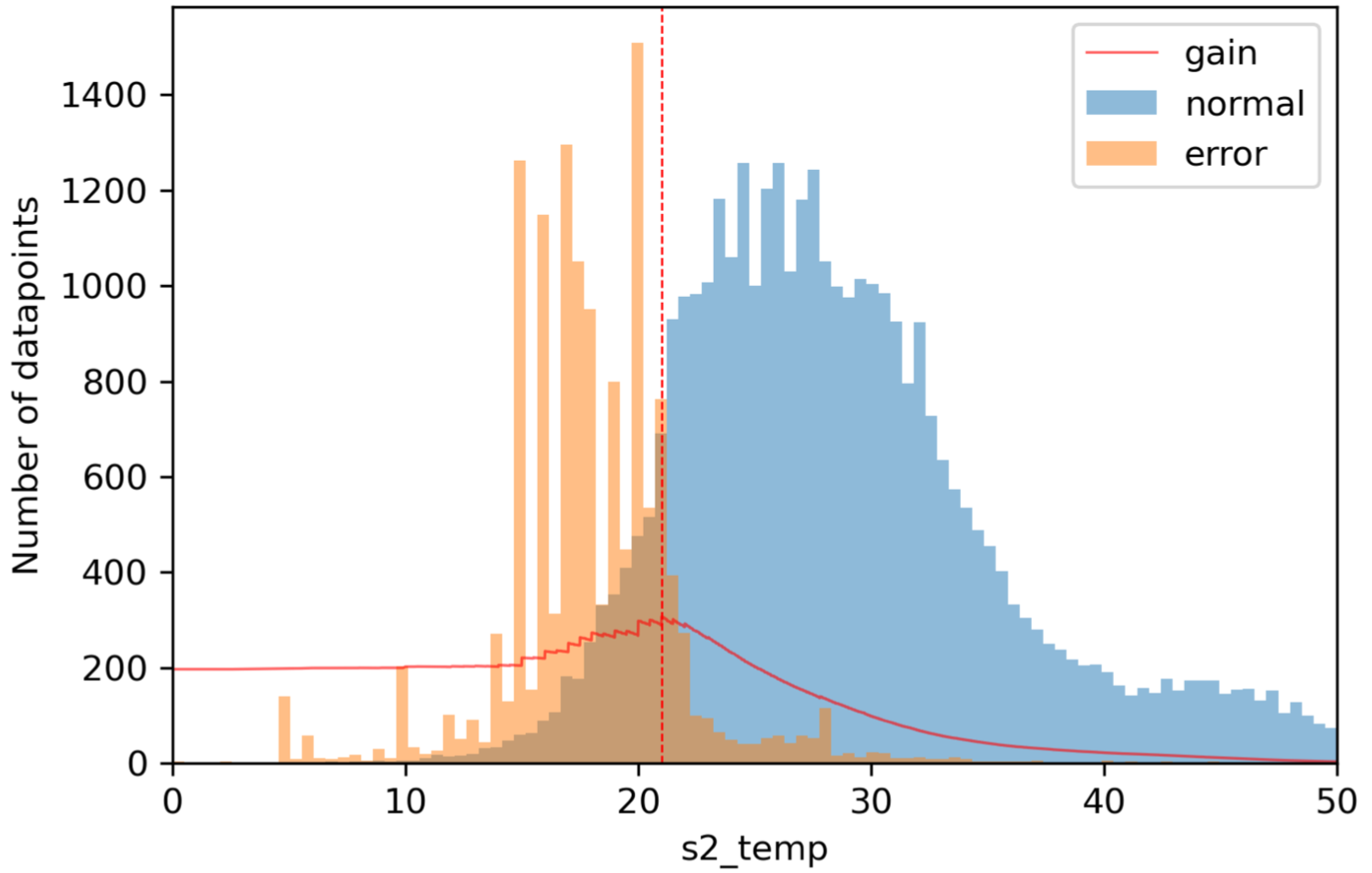}
    \caption{\textbf{Threshold selection} for the daily minimum of the sensor variable s2\_temp, w.r.t.\ the failure variable ``Warning low t1''. The optimal threshold is $\theta=21$, corresponding to gain $\gain=0.60$ (gain values are plotted with a 500 factor). The threshold applies to the values of s2\_temp: a good threshold splits the normal and error histograms.}
    \label{fig:gain_example_s2_temp}
\end{figure}

To find the optimal threshold for a dataset $D=s\mid f$, our algorithm compares the gains of the possible thresholds in the range of $s$.
\Cref{fig:gain_example_s2_temp} illustrates the selection of a threshold for the daily minimum of the sensor variable s2\_temp w.r.t.\ the failure variable ``Warning low t1''.
These variables are correlated only to a certain degree, as shown by the overlapping histograms:
the blue histogram counts the number of min s2\_temp values on days with normal heater operation (w.r.t.\ the failure ``Warning low t1'');
the orange histogram are values on days when that failure was recorded.
Clearly, normal values are related to higher temperatures, while the temperature measured on days with failures are comparatively lower.
The gain of each possible threshold is plotted in \Cref{fig:gain_example_s2_temp} as a red line at the bottom.
The sensor value that maximises the gain is the optimal threshold that our method learns, in this case \Celsius[21.0], shown as a red dashed vertical line.

Besides the optimal threshold, the algorithm returns the side of the split with the highest correlation to the failure class.
This discretises the sensor variable $s$ into a Boolean, with values $0$ on the split (left or right) with a majority of normal readings, and $1$ on the split with a majority of failure readings.

In \Cref{fig:gain_example_s2_temp}, a fault tree for the failure variable ``Warning low t1'' might use as basic or intermediate event the Boolean condition $\min(\text{s2\_temp})\leqslant 21$, associated to that system failure.

\subsection{Learning the fault tree}
\label{sec:method:learning}

In the second step of our approach, we use the Boolean data obtained during threshold learning, to generate and evaluate an FT for each failure mode.
Each input dataset contains all sensor variables (in one of their daily statistics) and one failure variable. 

\begin{figure}[htb]
    \centering
    \includegraphics[width=\linewidth]{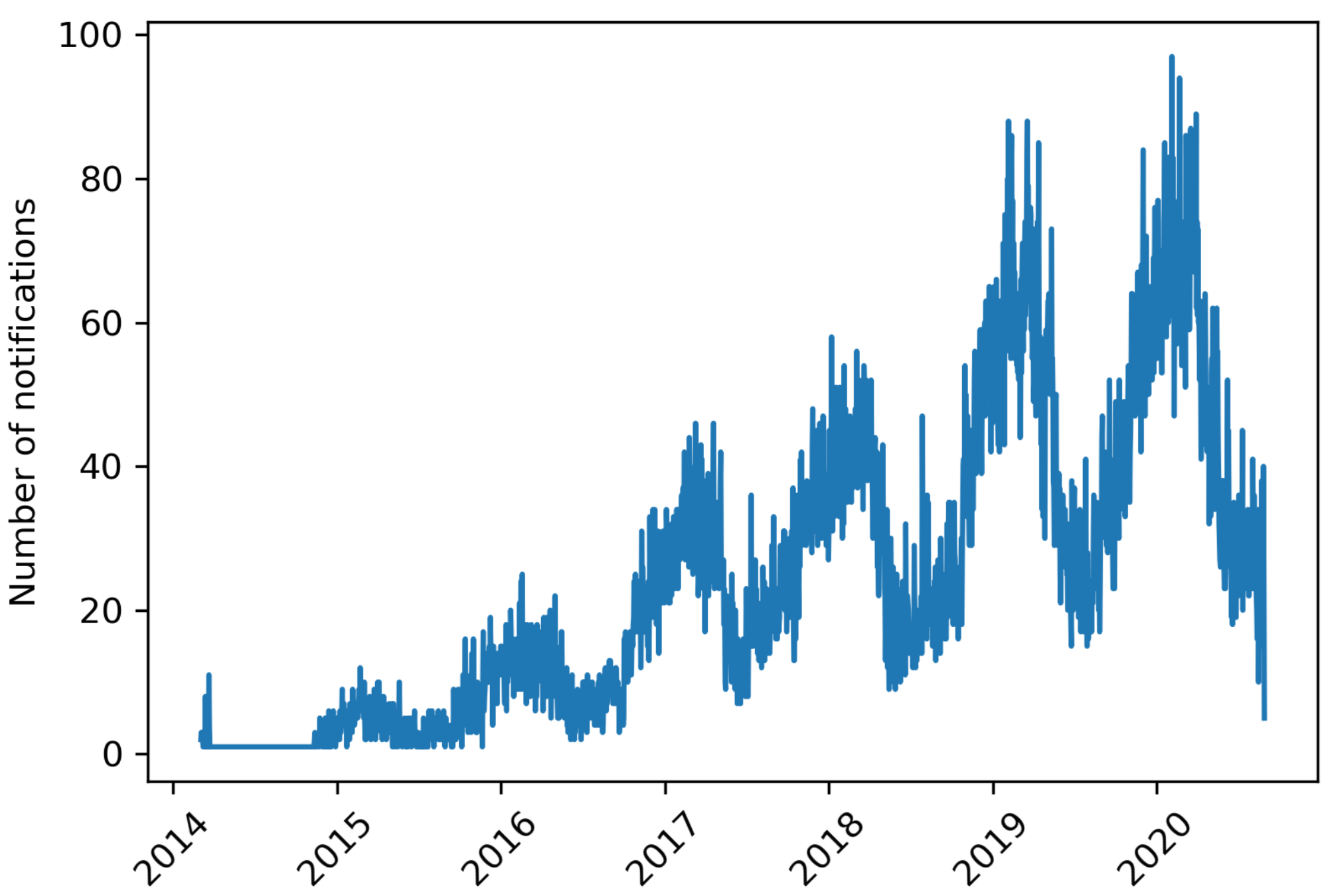} 
    %
    \caption{\textbf{Daily notifications} for ``Warning low ch\_pressure''}
    \label{fig:notifications}
\end{figure}

As an example of data availability for the failure variable, \Cref{fig:notifications} shows the number of notifications per day for ``Warning low ch\_pressure''.
The rising trend is due to the increasing number of heaters installed---c.f.\ \Cref{fig:data_collection}.
Seasonality is clear, as significantly more notifications are sent in months with lower outside temperatures in The Netherlands. For this failure mode, the dataset contains $52489$ daily data points with failure notifications. 

To each such data point showing failure, we add as counterpart a data point that describes a \textit{normal} operation regime. More than one choice may be valid for this. We select the most recent data point showing normal operation for the same heater. The nearest future datapoint after the heater is fixed would be another valid option, but options that take into account a longer history of operation for the heater are less valid, because the heaters have different life spans (or ages) in the dataset. Adding this counterpart leads to a combined, balanced dataset of roughly twice the size. This is the dataset used to generate the fault tree.

To learn the FT we use an improved version of our previous algorithm LIFT \cite{NBS18}.
Given a Boolean failure variable and the set of all real-valued sensor variables, thresholds are first learnt for the discretisation of the sensor variables into Boolean variables---henceforth referred to as \emph{thresholded variables}---as described in \Cref{sec:method:thresholds}. 

\begin{figure}[ht]
    \centering
    \includegraphics[width=.75\columnwidth]{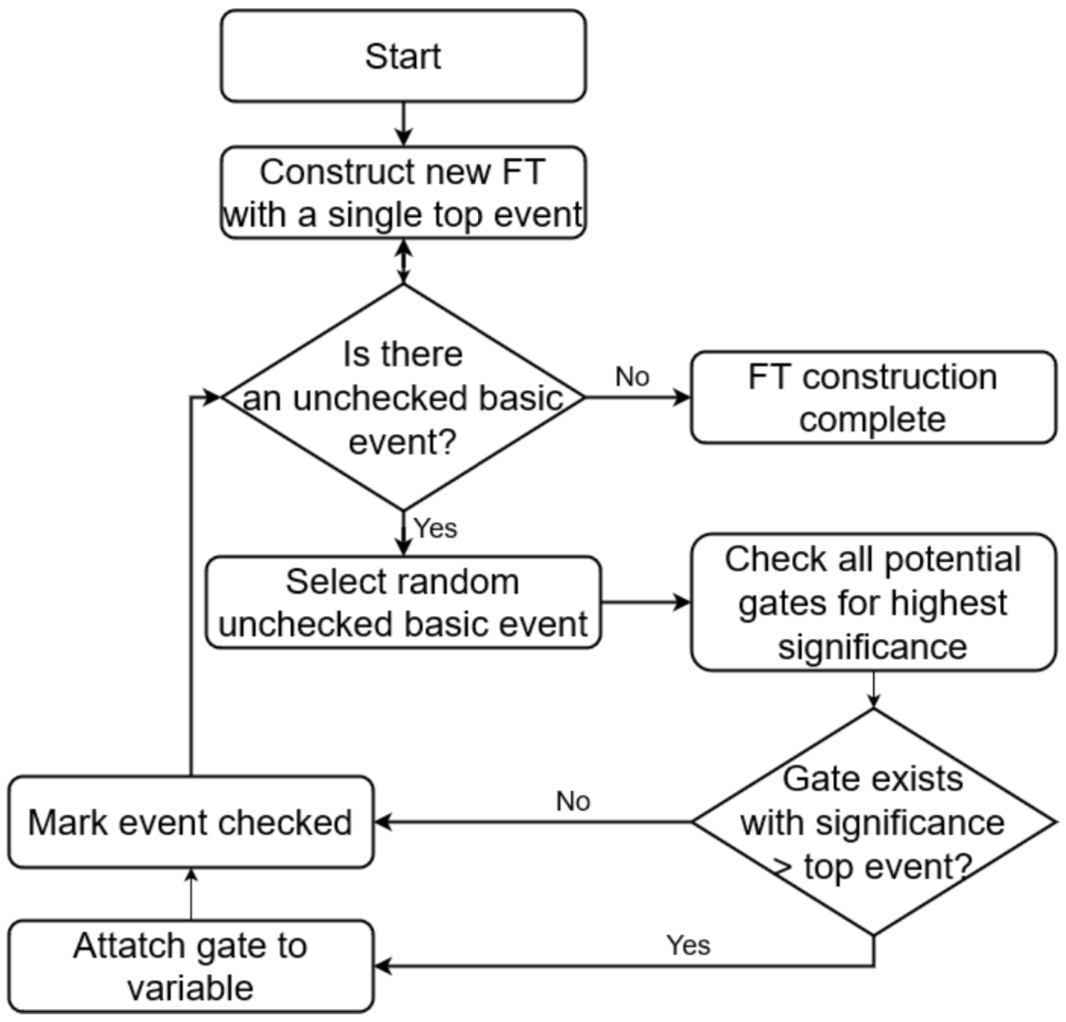} 
    %
    \caption{\textbf{The learning algorithm}}
    \label{fig:flowchart}
\end{figure}

We then initialise a fault tree $T$ with only the failure variable as TLE, and proceed iteratively as shown in the flowchart in \Cref{fig:flowchart}.
At each iteration, the algorithm chooses a thresholded variable that is not yet the output of a gate in $T$, and greedily searches for the best possible gate that can be constructed.
A logical gate is defined by its type (e.g.\ AND) and input variables.
LIFT explores all combinations of gate types and subsets of thresholded variables, up to a configured maximum number, set to 3 for our experiments.

\paragraph{The significance of a gate.}
During search, the gate with maximum \emph{significance} is chosen.
Intuitively, the significance of a gate measures the positive correlation between the output and the inputs of a gate.
The original implementation of LIFT used the Mantel-Haenszel statistical test for this \cite{NBS18}.
Instead, here we use the \emph{phi coefficient} as significance score \cite{Udn12,Cra46}.
It is a correlation coefficient for binary variables, whose value is 1 if the two variables are in complete agreement about their values, 0 indicates no relationship, and -1 complete disagreement.
In particular, the phi coefficient remains stable when the size of the dataset changes, unlike the Mantel-Haenszel statistical test.

For an illustration consider \textit{(i)} the failure variable ``Lockout code 11'', \textit{(ii)} the statistic that measures the range of values of a sensor variable in a day, and \textit{(iii)} the proposed gate:\\
\begin{minipage}{\linewidth}
\[
	\text{AND}(\text{s1\_temp}> 88.18 \,\boldsymbol,\, \text{pump\_pwm}> 0.0).
\]
\end{minipage}\\[1ex]
The significance measures the positive association of two variables:
{\fontfamily{cmr}\bfseries\selectfont1.}\ the output of the gate, i.e.\ the value of ``Lockout code 11''; and
{\fontfamily{cmr}\bfseries\selectfont2.}\ the Boolean expression, i.e.\ the AND gate over the thresholded sensor ranges ``s1\_temp'' and ``pump\_pwn''.

Additionally, we implement the following rules:
\begin{itemize}
\item	A new gate is added to the FT only if its significance
		is higher than or equal to that of the top gate.
\item	To limit the search space, each sensor variable is used
		at most in one location in the fault tree.
\end{itemize}
The purpose of these rules is to reduce the search space, thus lowering the runtime of LIFT.
This is also the reason why the maximum number of inputs chosen for the gates was set to 3 for our experiments.
Therefore, removing these rules can only increase the quality of the resulting FTs---at the cost of increasing the computation time.

The construction of an FT terminates when no gate can be added that increases its significance, or when all sensor variables have already been used.
The resulting fault tree $T$ is then reported together with the significance of its TLE.

Note that, since the algorithm only adds gates to $T$ that have higher significance than its TLE, the TLE significance is \emph{a lower bound of correlation} between gate inputs and output across all the gates in $T$.
Also, we annotate the BEs of $T$ with the probability $p$ of them occurring in the dataset---but, as indicated above, this probability is relative to a dataset which has been balanced between failure and normal operations.
As such, $p$ should only be interpreted in this context.

\subsection{Computational complexity and execution runtime}
\label{sec:method:complexity}

To prepare the data for the learning procedure, the Intergas database (terabytes of raw data) was processed in its Apache Spark framework for big data, executed in a large computing cluster.
This resulted in datasets prepared for FT learning, where the size of the resulting dataset was on the order of $10^5$ records per failure mode and daily statistic.
For these datasets, our learning procedure could be executed in a single computer, constructing each FT in a matter of minutes.

We highlight however that the procedure has high complexity: 
Namely, to find the optimal threshold during the discretisation of each sensor variable, all its unique values present in the data are considered.
This has linear complexity in the dataset size, but must be repeated for each of the 44 combinations of failure variables and daily statistics for which different thresholds are necessary.
Therefore, the aforementioned $10^5$ records still lead to relatively long execution runtimes.

Furthermore, LIFT considers all feasible gate combinations: a combinatorial problem in the number of the sensor and failure variables.
We cap this combinatorial explosion by limiting the maximum combination size to 3. 
Still, the resulting polynomial of degree 3 in the total number of variables causes the minutes (rather than seconds) runtimes mentioned above.

\section{Results}
\label{sec:results}

We obtain 44 fault trees with various significance levels. We first present the fault trees with the highest significance, then provide a summary of results. We note that, for this case study, from among the four basic statistics of the sensor data, the daily minimum, maximum, and range tended to be the most useful, leading to fault trees with higher significance than the daily average. 

\begin{figure}[htb]
    \centering
	\includegraphics[scale=.7]{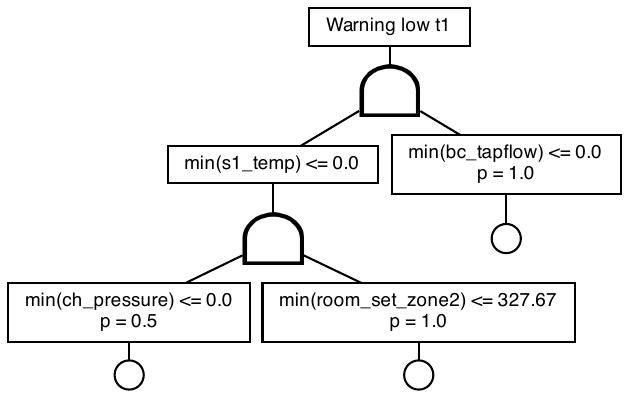}
    \caption{\textbf{Fault tree} for ``Warning low t1''. Significance 0.96.}
    \label{fig:FT-Wlt1-min}
\end{figure}

\paragraph{Fault tree for ``Warning low t1''.} This failure mode makes for a good test of the fault-tree learning method, because it has a clear semantic link to the s1\_temp sensor variable for the supply water temperature (which is thus expected to appear in the fault tree). This is indeed what we obtain: Fig.~\ref{fig:FT-Wlt1-min} shows the fault tree (with nearly maximum significance, 0.96) for this failure mode. When the daily statistic min(s1\_temp) falls at or below $\theta = 0$, this occurrence strongly associates with the failure ``Warning low t1'', with gain 0.86. This is the main reason of the failure notification, but the tree provides more information. An interruption in the water supply, min(bc\_tapflow) $\leqslant 0.0$, is also linked to the failure. Then, low minimum daily readings for s1\_temp associate with a temporary lack of water pressure, min(ch\_pressure) $\leqslant 0.0$. The final sensor variable, room\_set\_zone2, whose value is always below the threshold ($p=1\%$), is due to the gate syntax requiring more than one input, and can be omitted. 

\begin{figure}[htb]
    \centering
    \includegraphics[scale=.7]{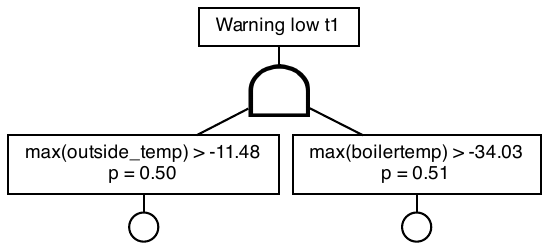}
    \caption{\textbf{Fault tree} for ``Warning low t1''. Significance 0.95.}
    \label{fig:FT-Wlt1-max}
\end{figure}

An alternative fault tree (also with nearly maximum significance, 0.95) for this failure mode is in Fig.~\ref{fig:FT-Wlt1-max}. This uses maximum daily readings and is less intuitive, since the maximum daily value of s1\_temp are not relevant for the failure, the way the minimum daily value is. Instead, two other daily maximums are significantly associated with ``Warning low t1'': the sensor variables boilertemp (whose maximum daily values above $\theta = -34.03$ are strongly associated with the failure ``Warning low t1'', with gain 0.81) and outside\_temp (similarly, above $\theta = -11.48$ with gain 0.84). These extremely low temperature thresholds are explained by the behaviour of the temperature sensor: it records in the data a very broad range [-51, MAX] \Celsius of maximum temperature values, where MAX is the maximum possible stored value on the sensor register (327.67 \Celsius). For the two sensors present in this fault tree (boilertemp and outside\_temp), the MAX temperature occurs in the data much more often than realistic temperatures, and thus must signal a local sensor failure. When both of these sensors record extreme maximum daily values, this associates with the failure mode ``Warning low t1'', and could thus be used as an early warning sign.

\begin{figure}[htb]
    \centering
    \includegraphics[scale=.7]{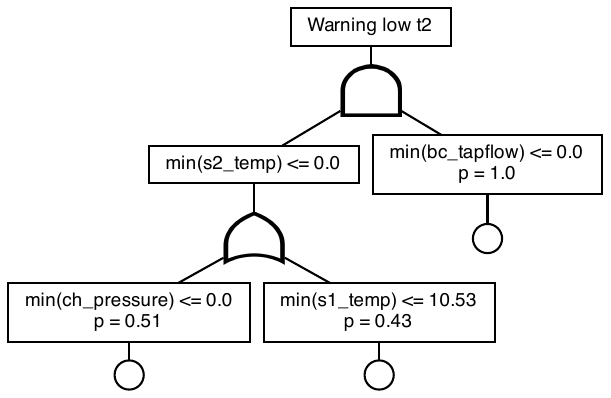}
    \caption{\textbf{Fault tree} for ``Warning low t2''. Significance 0.96.}
    \label{fig:FT-Wlt2-min}
\end{figure}

\paragraph{Fault tree for ``Warning low t2''.} This failure variable is similar to ``Warning low t1'' in that it has a semantic link to a temperature sensor variable, here s2\_temp. In the resulting fault tree shown in Fig.~\ref{fig:FT-Wlt2-min}, s2\_temp appears as expected, with minimum daily values at or below $\theta = 0$ strongly associated with ``Warning low t2'', gain 0.87. In this case though, the fault tree also shows a different intuitive association: a temporary lack of water pressure (ch\_pressure) {\tt or} a low supply water temperature (s1\_temp) associate with a low return water temperature (s2\_temp). It is also likely that this is a causal, not only correlational, relationship between these sensor variables, since the water supply system precedes (and must affect) the return water system.

\begin{figure}[htb]
    \centering
    \includegraphics[scale=.7]{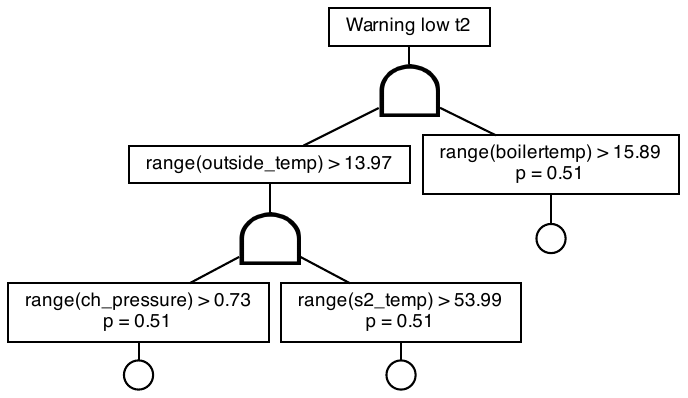}
    \caption{\textbf{Fault tree} for ``Warning low t2''. Significance 0.95.}
    \label{fig:FT-Wlt2-range}
\end{figure}

Also here, an alternative fault tree (significance 0.95) for this failure mode is in Fig.~\ref{fig:FT-Wlt2-range}, now using the daily range statistic. ``Warning low t2'' associates with days with large variation in  outside temperature, range(outside\_temp) $>13.97$ (gain 0.85), and the sensed machine temperature range(boilertemp) (gain 0.84). For the sensor outside\_temp, the range readings are above 0 and mostly below 80 \Celsius. This range of outside\_temp itself associates with the range of ch\_pressure (gain 0.75) and that of s2\_temp (gain 0.73), sensor variables which were also found predictive in the FT from Fig.~\ref{fig:FT-Wlt2-min} using the minimum daily readings. We note that the daily minimum of s2\_temp is more predictive of ``Warning low t2'' than the range of s2\_temp: this variable appears higher in the FT using daily minimums. However, overall, both minimum and range daily readings are similarly predictive for this failure, with small and significant FTs obtained.

\paragraph{Fault tree for ``Lockout code 11''.} The fault trees obtained so far for temperature-related failure modes had nearly maximum significance levels, suggesting the presence of strong associations among the variables. This is not always the case: other failure modes are more complex, and much harder to model out of noisy data, resulting in lower significance levels and deeper fault trees. 

\begin{figure*}[bth]
    \centering
    \includegraphics[scale=.6]{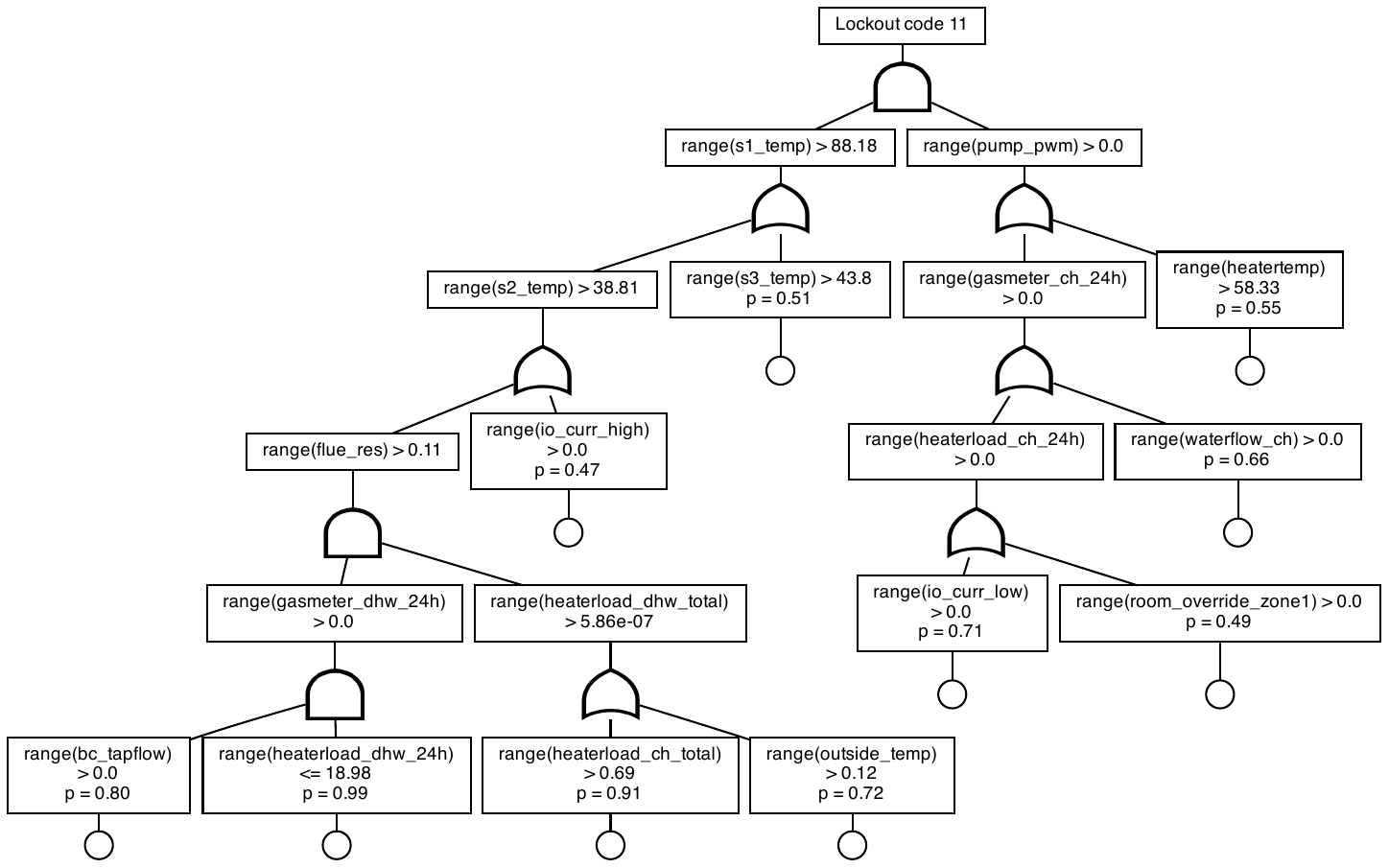}
    \caption{\textbf{Fault tree} for ``Lockout code 11''. Significance 0.35.}
    \label{fig:FT-Lc11-range}
\end{figure*}

For the failure mode ``Lockout code 11'' (a sensor fault also related to s1\_temp), the fault tree with the highest significance is shown in Fig.~\ref{fig:FT-Lc11-range}. This tree uses the daily range statistic of the sensor variables. It has a significance of 0.35, meaning that, at every gate, the input-to-output correlation is at least 0.35. In other words, only part of the reasons for failure are learnt out of the data available, and either more exist but are not sensed in the data, or the failure is partly random. We describe below the associations which were learnt.

The left-hand side of the tree suggests that the failure can be predicted by the behaviour of the water supply temperature s1\_temp, although only partly (gain 0.09). The failure correlates with a very large daily range for s1\_temp: this range varies between 0 and 180.2 \Celsius in the data (a large range, due to the presence of extreme temperature readings, as described for ``Warning low t1''). The higher the daily range value, the more likely the failure is. In turn, range(s1\_temp) is correlated with the ranges of the other two temperature sensors s2\_temp and s3\_temp, whose values vary mostly between 0 and 100 \Celsius in the data. The range of the return water temperature s2\_temp is then correlated with (and possibly caused by) the range of the current on high output power io\_curr\_high and the resistance of the flue duct flue\_res, with higher values always signalling failure. The right-hand side of the tree states that above-zero control signal for the modulating pump (range(pump\_pwm), with values in [0, 100] \%) is also a predicting factor for this failure mode, although also weakly (gain 0.03). A combination of other technical variables are further predictive for range(pump\_pwm).

\begin{figure}[htb]
    \centering
    \includegraphics[scale=.7]{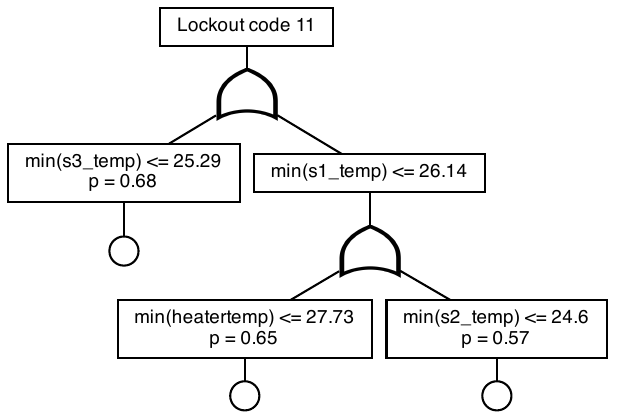}
    \caption{\textbf{Fault tree} for ``Lockout code 11''. Significance 0.27.}
    \label{fig:FT-Lc11-min}
\end{figure}

An alternative FT for ``Lockout code 11'', obtained using minimum daily sensor data, is shown in \Cref{fig:FT-Lc11-min}. It is much smaller and slightly less significant (significance 0.27). The minimum daily readings for four temperature sensors (s1\_temp to s3\_temp and heatertemp) are slightly predictive of this failure: in all cases, if the minimum daily temperature falls below a threshold of 24-27 \Celsius, the failure is more likely to occur.

Such alternative fault trees learnt a partial pattern for the failure from different sets of daily statistics. They can then be used together, as a predictive ensemble for the failure mode.

\begin{figure}[ht]
    \centering
    \includegraphics[scale=.6]{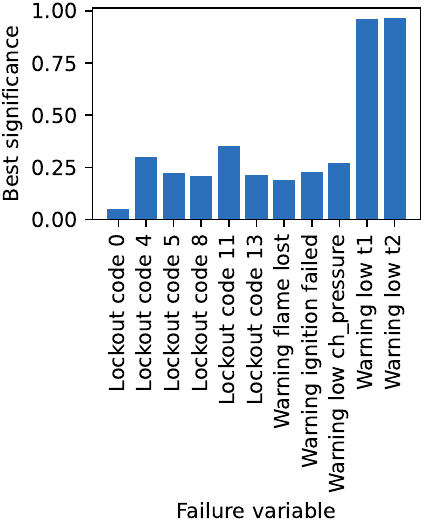}
	\qquad
	\includegraphics[scale=.6]{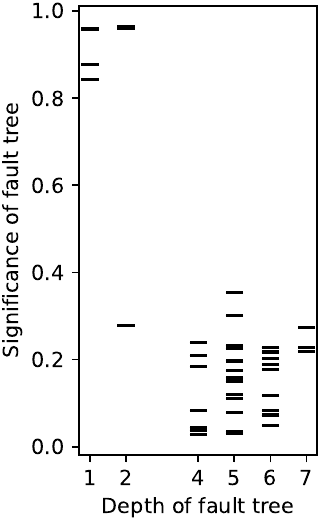}
    \caption{\textbf{Summary of results}: significance scores (left) and significance vs.\ depth for all fault trees obtained (right).}
    \label{fig:summary}
	\vspace{-1ex}
\end{figure}

\begin{figure*}[htb]
    \centering
    \includegraphics[scale=.6]{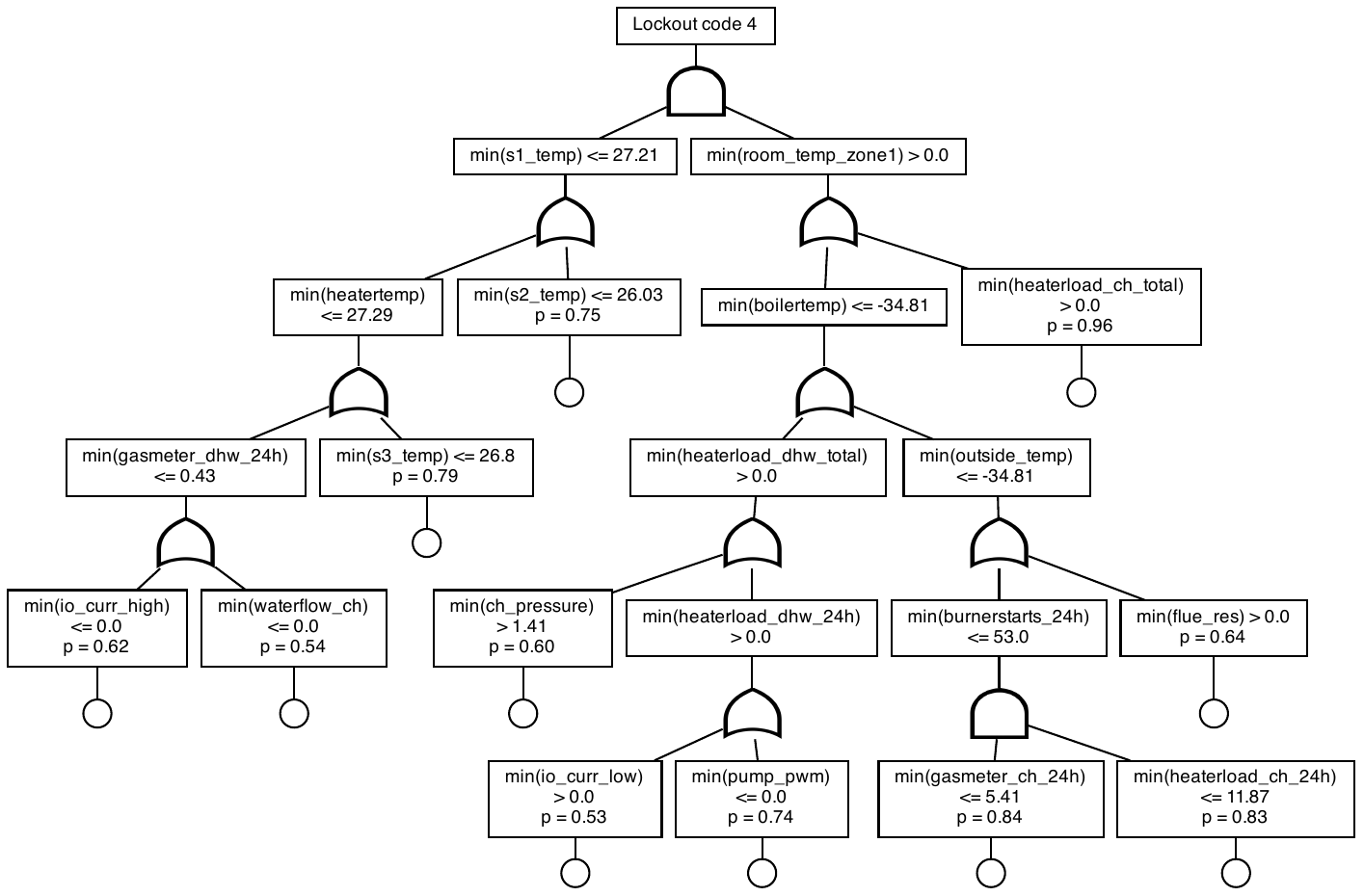}
    \caption{\textbf{Fault tree} for ``Lockout code 4''. Significance 0.30.}
    \label{fig:FT-Lc4-min}
\end{figure*}

\paragraph{Fault tree for ``Lockout code 4''.} This failure mode triggers when there is no flame signal, so it is of a different nature than the previous examples. However, the set of predictive sensor readings and their thresholds are similar to those which also predict other failures, so the failures themselves are likely dependant. The best fault tree obtained for ``Lockout code 4'' is shown in \Cref{fig:FT-Lc4-min} (significance 0.30) and uses minimum daily readings. Some temperature sensors (s1\_temp to s3\_temp) appear with similar thresholds as for ``Lockout code 11'' (the fault tree in \Cref{fig:FT-Lc11-min}), and they are the sensor variables with the highest gain, so strongest individual association to the failure. In total, 20 out of 27 sensor variables available are thresholded and included, leading to a complex fault tree.

\paragraph{Summary of results and evaluation.}
We obtained 44 FTs of various depths, internal structure, and significance scores: \Cref{fig:summary} provides a quantitative overview. Trees with low depths tend to have high significance. With regards to the best significance score per failure variable, we observe three classes:
(a) failure variables for which the significance score was nearly maximum (``Warning low t1'' and ``\ldots t2''),
(b) those for which the significance is medium-to-low, between 0.21 and 0.35 (the majority of the cases), and
(c) those for which no significant FT could be learnt (one case: Lockout code 0).
The depths of the trees, defined as the number of gates on the longest path from the TLE to a BE, varies between 1 and 7.
For our case study, we find that the FTs are deep only when they also have low significance.

The runtime of the method to construct one fault tree, for this dataset and algorithm configuration, is a few minutes (for threshold computation) and a few tens of seconds (for the tree generation) on an average consumer desktop computer.

\section{Discussion and Conclusions}
\label{sec:conclu}

\paragraph{Evaluation.}
To evaluate the FTs learnt, we consulted with Intergas domain specialists, who confirmed that the fault trees do contain meaningful relationships between variables.
In cases where the domain experts already had good understanding of how a failure occurs, they also confirmed that the important variables were in the tree.

\paragraph{Limitations.}
Many design decisions may affect the accuracy of the resulting fault trees.
In particular we note:
\begin{description}
    \item[High runtimes] The method has high computational complexity, although it can be configured so that it remains feasible. Our runtime per fault tree with a dataset on the order of $10^5$ records and 27 sensor variables per record is on the order of minutes in a standard desktop computer.
    \item[Lack of optimality] This learning procedure, as many classical machine-learning algorithms such as those based on decision trees~\cite{BFOS84} is a greedy heuristic: while the gate with the highest significance available is constructed at every step, the average significance score over the entire fault tree may not be maximum. As a consequence, there may be other good alternatives for the tree structure. 
    \item[Single threshold: bias is likely] Model bias can stem from making oversimplified assumptions about the data. We compute only one threshold per sensor variable (a potential cause of bias), and then use the result at most once in an FT. However, in some datasets, multiple thresholds for the same variable may be of interest, in different parts of the tree. We made the choice to disregard this option and precompute this single threshold per sensor variable and failure variable, to keep the runtimes feasible.
    \item[Real-world data is faulty and complex] We observed unordinary sensor values (very low, very high, and zero), particularly from temperature sensors; these are documented in the sensor data sheets \cite{Ban21}, and are unavoidable during the lifetime of a sensor. Both noise and unusual or missing (zero) values are pervasive in the data: we have only cleaned the dataset minimally. The resulting FTs have learnt from these unusual values.
    \item[Overfitting to noise is possible] In datasets where noise is frequent, it is possible that some of the low-significance trees overfit (learn from) noise.
    \item[No proven causality] The fault trees learn correlations rather than causal relationships, which are difficult to extract from passive (non-interventional) datasets like ours.
    \item[No cover for dynamic failures] We use the LIFT algorithm, which can learn static fault trees (AND and OR gates). However, some failures may be dynamic in nature, e.g.\ when failure order matters. This cannot be captured by static gates, and poses a much harder learning problem.
\end{description}

We note, nonetheless, that some of these limitations could be resolved in other case studies.
We designed and configured the method to keep the runtime low, but some of the restrictions can be lifted if the runtime remains feasible.

\paragraph{Discussion of results.}
We were able to address our research questions to different degrees of success.

Regarding \ref{RQ:howto}, the combination (and modification) of the C4.5 and LIFT algorithms allowed us to build FTs automatically from sensor data.
The basic and intermediate events of these trees were chosen based on the correlations of failure behaviour and sensor values, which was our main objective.

Access to failure variables was instrumental to achieve automation, since these guide both steps of our learning algorithm.
More precisely, for each FT: 
(1) we discretise all sensor values, by learning failure behaviour based on the available failure variables, and
(2) we evaluate all possible combinations of FT structures, selecting a failure variable as TLE.

As mentioned above, our decision to discretise each sensor variable with a single threshold may be imposing unneeded limitations.
Still, some resulting FTs achieved high quality, both according to our metrics and by expert assessment.

In that respect and regarding \ref{RQ:howgood}, we use gate significance---lifted to FTs by taking the significance of its TLE---as metric for the the quality of our trees.
The original implementation of LIFT used a similar concept, which we modified as indicated in \Cref{sec:method:learning} to cater for variable dataset sizes.
An extra advantage of our choice, for this case study, is that larger FTs had generally lower significance.
Smaller FTs are easier to interpret by humans, and thus preferable for risk management.

Finally, \ref{RQ:big} was answered in a positive manner, as our implementation could handle the input of a company with several millions of data readings.
Although the initial data processing and filtering was performed in the cluster of Intergas, this is not considered a limitation since data protection regulations make it a usual case.
More relevant are our rules to keep the learning problem within manageable size for a desktop computer.
We achieved our goal, but loosing restrictions to favour the FT quality---e.g.\ allowing several thresholds per sensor variable---will have a direct impact in the execution runtime of the core learning algorithm.

\paragraph{Perspectives and future work.}
A first point of improvement would be to allow repeated uses of a sensor variable in an FT, with different thresholds, and a higher number of children per gate.
The challenge lies in keeping the runtime reasonable, which may call for rules such as scoring and penalising sensor variables already present in an FT.
Semi-automatic FT construction is promising for this, to mitigate the combinatorial explosion faced by the FT learning step.
We intend to contribute to public benchmarks such as \cite{RBN+19}, by submitting the fault trees of our work.

\section*{Acknowledgements}

Funded by the European Union under GA number 101067199 ProSVED.
Views and opinions expressed are those of the author(s) only and do not necessarily reflect those of the European Union or The European Research Executive Agency.
Neither the European Union nor the granting authority can be held responsible for them.


\bibliographystyle{apacite}
\PHMbibliography{main.bib}  



\end{document}